\documentclass{article} 
\usepackage{iclr2027_conference,times}

\usepackage{amsmath,amsfonts,bm}

\def\eqref#1{equation~\ref{#1}}

\def\1{\bm{1}}

\DeclareMathAlphabet{\mathsfit}{\encodingdefault}{\sfdefault}{m}{sl}
\SetMathAlphabet{\mathsfit}{bold}{\encodingdefault}{\sfdefault}{bx}{n}

\usepackage{hyperref}
\usepackage{url}

\usepackage{amsmath} 
\usepackage{caption}
\usepackage{subcaption} 
\usepackage{enumitem}
\usepackage{multirow}
\usepackage{cleveref}
\usepackage{algorithm}
\usepackage{algpseudocode}
\usepackage{graphicx} 
\usepackage{amssymb}
\usepackage{booktabs}

\usepackage{colortbl}
\definecolor{lightblue}{rgb}{0.85,0.92,1.0} 
\newcommand{\MethodName}{WorldRecon}

\title{\MethodName{}: Generative Multi-View Mesh Reconstruction for Large-Scale 3D World}

\author{
\textbf{Mengchen Zhang}$^{1,2}$,
\textbf{Yang Li}$^{2}$,
\textbf{Chunshi Wang}$^{1,2}$,
\textbf{Tong Wu}$^{3}$,
\textbf{Chunchao Guo}$^{2}$ \\[0.6em]
\makebox[\textwidth][c]{%
\normalfont
\begin{tabular}{c}
  $^{1}$ Zhejiang University \qquad
  $^{2}$ Tencent Hunyuan \qquad
  $^{3}$ Fudan University \qquad
\end{tabular}%
}
}

\iclrfinalcopy 
\begin{document}

\maketitle
\lhead{} 
\renewcommand{\headrulewidth}{0pt}
  
\begin{center}
    \centering
    \vspace{-10pt}
    \includegraphics[width=\linewidth]{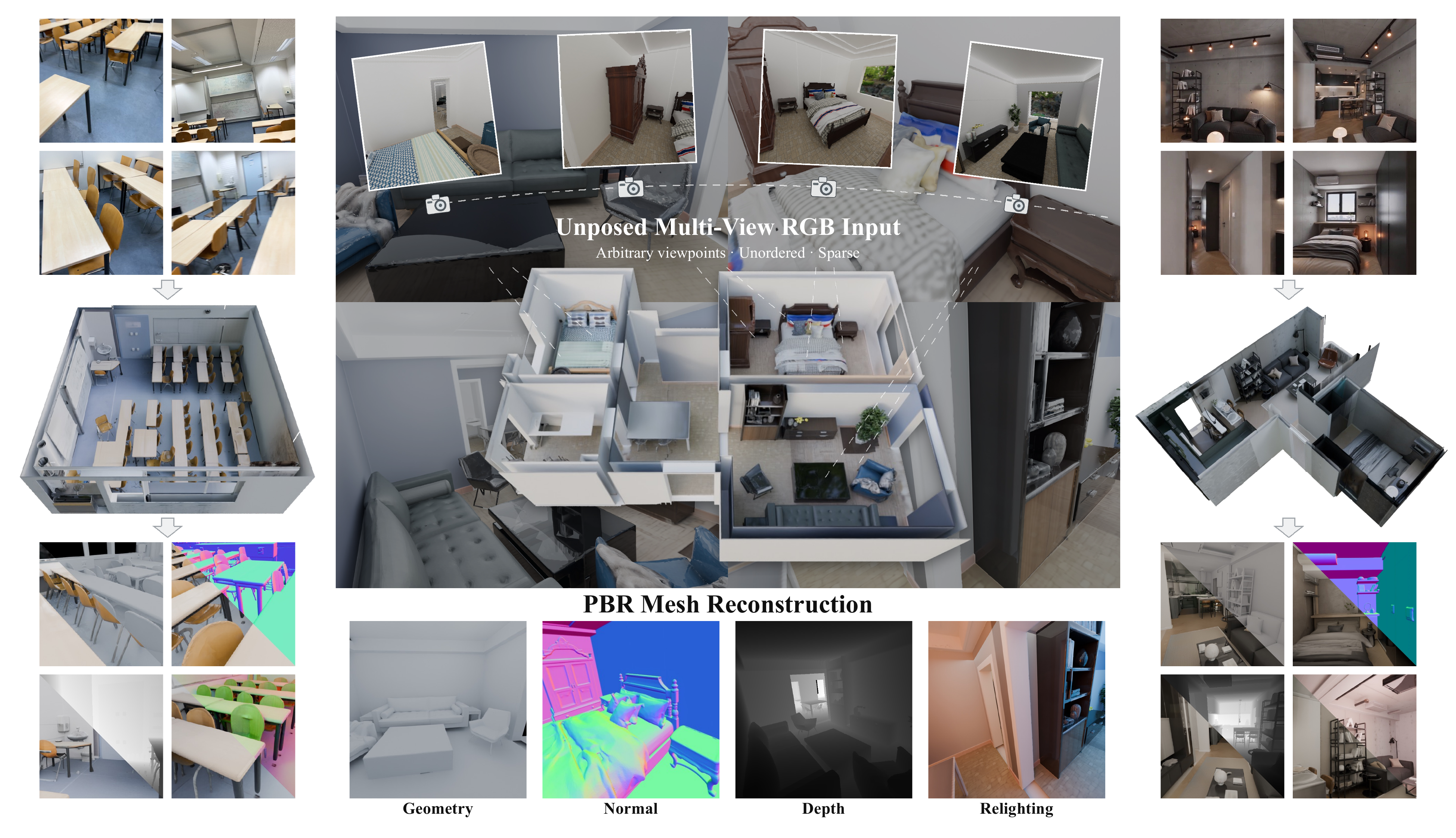}
    \vspace{-20pt}
    \captionof{figure}{\textbf{\MethodName{} Overview.} Given unposed multi-view RGB images of an indoor scene, \MethodName{} reconstructs a complete scene-scale PBR mesh block-by-block via a native 3D generative prior at arbitrary scene scales, yielding high-fidelity geometry, depth, normal, and texture. From left to right, we demonstrate \MethodName{} on real-world captures, synthetic images, and frames from generated videos, highlighting generalization across diverse input domains.
    }
	\label{fig:teaser}
\end{center}

\begin{abstract}
Reconstructing complete, high-quality meshes of large-scale scenes from unposed multi-view images requires faithful geometric grounding while plausibly completing surfaces that are unobserved or only sparsely captured. We present \MethodName{}, a generative multi-view reconstruction framework for recovering complete, large-scale 3D scene meshes from unposed RGB images. \MethodName{} uses a coarse, noisy global point cloud and camera poses estimated from the input images by a reconstruction model as a global scaffold for whole-scene generation and partitions it into overlapping spatial blocks. To robustly leverage these imperfect geometric priors, we equip the native 3D generative model TRELLIS.2 with a dedicated reconstruction-based conditioning architecture that generates each block from its local point-cloud crop and block-relevant image observations, producing sparse structure, detailed geometry, and PBR materials. To maintain scene-level coherence, we introduce neighbour-aware latent inpainting and a causal world-generation procedure that propagates structural, geometric, and material context across overlapping blocks. Together, these designs scale multi-view generative reconstruction to large-scale scenes, yielding coherent, unified meshes. Extensive experiments demonstrate that \MethodName{} achieves state-of-the-art reconstruction performance, reducing Chamfer distance by \(14.9\%\) and \(43.1\%\) while improving F-score by \(17.3\%\) and \(12.0\%\) over the strongest baselines on synthetic and real-world indoor scenes, respectively, thereby demonstrating faithful and complete generative reconstruction as well as strong sim-to-real generalization. Project page:   \url{https://kszpxxzmc.github.io/WorldRecon-website/}.

\end{abstract}

\section{Introduction}
\label{sec:Introduction}

Recovering a large-scale scene from unposed RGB images is a fundamental problem in computer vision, with broad applications in virtual and augmented reality, embodied intelligence, and digital content production. Reconstruction-based methods~\cite{DUSt3R,MASt3R,VGGT-Omega,Mip-nerf360,MonoSDF,3DGS} recover scene geometry from visual observations, but often leave occluded regions incomplete and produce noisy geometry that is difficult to convert into clean meshes with detailed surfaces and materials.

Generative priors improve reconstruction completeness by inferring plausible geometry and appearance beyond the observations. Building on native 3D generative models~\cite{TRELLIS,TRELLIS2,CLAY}, recent multi-view extensions~\cite{ReconViaGen,UniRecGen} directly reconstruct structured 3D assets from unposed RGB images. Extending these methods to scene scale, however, remains challenging as the spatial extent and geometric complexity of an environment exceed the capacity of a single high-resolution generative representation.

Recent scene generation methods~\cite{TRELLISWorld,BlockFusion,LT3SD,WorldGrow,SynCity} scale to substantially larger environments by decomposing a scene into overlapping spatial blocks and generating them jointly or progressively. However, they are typically conditioned on text or layouts and thus lack visual guidance and 3D spatial grounding. More closely related to our task, scene-level generative reconstruction methods represent large scenes using hierarchical voxel latents~\cite{SCube} or overlapping scene chunks~\cite{GenRecon}. However, they require accurate camera poses to project image features into a 3D feature volume, making reconstruction quality sensitive to pose errors.

In this work, we introduce \MethodName{}, a generative multi-view framework for reconstructing large-scale scene meshes from unposed RGB images. We employ VGGT-$\Omega$~\cite{VGGT-Omega} to estimate coarse camera poses and recover a global point-cloud scaffold, which we partition into overlapping blocks. Our reconstruction-based conditioning leverages local point-cloud crops and block-relevant image observations, enabling the generation of complete, high-fidelity mesh blocks with PBR materials from imperfect geometric and camera estimates.
To promote consistency across blocks, we train neighbour-aware latent inpainting models that condition each new block on overlap latents from previously generated neighbours. At inference, we progressively expand the scene following a causal block order, propagating sparse structure, geometry, and material context through shared world representations. This enables neighbouring blocks to transition seamlessly in both geometry and appearance before being decoded and assembled into a unified scene mesh.
  
Extensive experiments on 3D-FRONT~\cite{3DFront} and ScanNet++~\cite{Scannet++} demonstrate that \MethodName{} reconstructs globally accurate and complete indoor scenes with fine geometric and material details from unposed RGB images on both synthetic and real-world settings.

In summary, our main contributions are as follows:
\begin{itemize}[leftmargin=*]
  \item We present \MethodName{}, a framework that integrates 3D reconstruction and generation to recover large-scale, high-fidelity scene meshes with PBR materials from unposed multi-view RGB images.
  
  \item We develop reconstruction-based block generation with neighbour-aware inpainting, producing faithful mesh blocks from local observations and imperfect geometric and camera estimates.
  
  \item We propose a causal world-generation pipeline that propagates sparse structure, geometry, and material context across overlapping blocks for scalable and seamless scene generation.
  
  \item Extensive experiments demonstrate state-of-the-art performance in large-scale mesh reconstruction from unposed multi-view images across synthetic and real-world indoor scenes.
  \end{itemize}

\section{Related Work}
\label{sec:Related_Work}

\noindent\textbf{Multi-View 3D Reconstruction.}
Traditional 3D reconstruction relies on Structure-from-Motion~\cite{SfM,GSfM} and Multi-View Stereo~\cite{MVS,PWMVS} to recover camera poses and scene geometry through triangulation. 
Learning-based MVS methods~\cite{MVSNet,R-MVSNet,Patchmatchnet} further improve quality and efficiency via differentiable cost volumes. 
More recently, pose-free models~\cite{DUSt3R,MASt3R,VGGT,VGGT-Omega} cast the pairwise reconstruction problem as a regression of pointmaps, jointly regressing cameras and dense geometry.
Beyond discrete point clouds, Neural Radiance Fields~\cite{NeRF,Ibrnet,Mip-nerf360} reframe reconstruction as inverse volumetric rendering for photorealistic view synthesis.
Neural implicit surface methods~\cite{NeuS,VolSDF,MonoSDF} represent scenes as continuous signed-distance functions.
3D Gaussian Splatting~\cite{3DGS,2DGS,pixelsplat} optimize scenes via differentiable rasterization for fast rendering. 
However, these methods suffer from incomplete reconstruction and tend to produce over‑smooth, blurred details in occluded regions. 

\noindent\textbf{Generative Multi-View 3D Reconstruction.}
Generative priors are introduced in multi-view 3D reconstruction to mitigate reconstructive incompleteness.
Early efforts~\cite{SyncDreamer, CAT3D, ReconFusion, Murre, GeometryCrafter}
leverage 2D or video diffusion priors to synthesize novel views or
depth maps, but still require per-scene optimization and suffer from cross-view inconsistency.
Other works~\cite{LVSM, LGM, LongLRM} regress target-view pixels or
3D Gaussians in a single feed-forward pass, but output unstructured primitives rather than meshes.
Recently, native 3D generative models~\cite{TRELLIS, TRELLIS2, CLAY} have enabled direct mesh generation from image conditioning.
ReconViaGen~\cite{ReconViaGen} and UniRecGen~\cite{UniRecGen} extend
this paradigm to multi-view conditioning, but remain confined to the
object level. SCube~\cite{SCube} and GenRecon~\cite{GenRecon} further extend multi-view 3D reconstruction to the scene level, yet both rely on accurate camera poses to project image features into a 3D feature volume, which induces cross-view conflicts and degrades reconstruction quality under imperfect poses.
In contrast, our method conditions on pose-free RGB images, recovering coherent scene mesh block-by-block via a native 3D generative prior.

\noindent\textbf{Unbounded 3D Scene Generation.}
Extending 3D generation from objects to unbounded scenes has recently drawn increasing attention. Early efforts focus on bounded environments such as a single room, either retrieving assets from a fixed database~\cite{DiffuScene, SceneFormer}, or hallucinating RGB-D frames with 2D diffusion models under text or layout conditions~\cite{Text2Room, ControlRoom3D}. 
To scale beyond bounded rooms, recent works pursue unbounded scene generation via block-wise generation, where neighbouring blocks are either jointly denoised and seamlessly blended~\cite{TRELLISWorld}, or progressively synthesized by an extrapolation prior~\cite{BlockFusion, LT3SD, WorldGrow, SynCity}.
However, these methods are either unconditional or conditioned only on text or layout, lacking both visual guidance and 3D spatial grounding. 
Our method instead grounds the scene in a coarse predicted point cloud and conditions each block on a few multi-view images, scaling multi-view generative reconstruction to arbitrarily large scenes.

\section{\MethodName{} Method}
\label{sec:Method}

\begin{figure}[t]
    \centering
    \includegraphics[width=\linewidth]{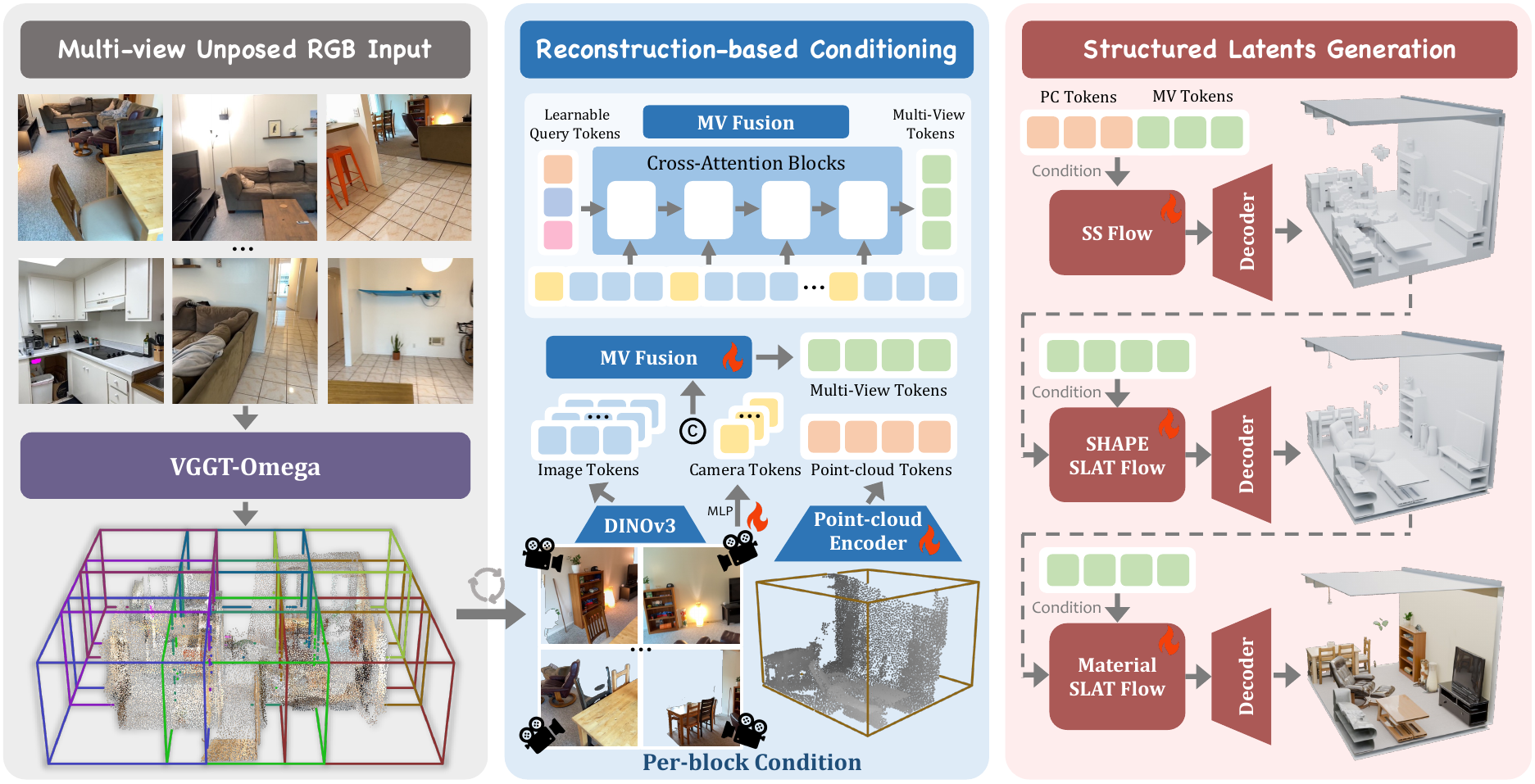}
    \caption{\textbf{\MethodName{} Pipeline.} 
      \textbf{Left:} Given unposed multi-view RGB images, VGGT-$\Omega$ reconstructs a global point cloud. The point cloud is divided into overlapping blocks, each paired with local RGBA images and predicted camera poses.
      \textbf{Middle:} For each block, DINOv3 image tokens and Pl\"ucker-ray
      camera tokens are fused into fixed-length multi-view tokens, while a
      point-cloud encoder extracts geometric tokens.
      \textbf{Right:} Three cascaded flow models progressively generate the
      sparse structure, detailed geometry, and PBR materials.
      }
	\label{fig:pipeline}
\end{figure}

Given an arbitrary number of uncalibrated RGB images $\{I_n\}_{n=1}^{N}$ of a large-scale indoor scene, our goal is to recover a complete, high-fidelity, scene-level 3D mesh endowed with PBR materials.
Our generative multi-view reconstruction framework couples a reconstruction prior for global layout consistency with a generative prior for surface fidelity and completeness. 
To this end, we curate a block-level training dataset (\Cref{sec:Method_Data}). We finetune the object-level generative prior of TRELLIS.2~\cite{TRELLIS2} into a block-level mesh generator conditioned on each block's point-cloud slice and multi-view RGBA images, together with a neighbour-aware 3D inpainting variant (\Cref{sec:Method_Block}). At inference, we first extract a global point cloud with VGGT-$\Omega$~\cite{VGGT-Omega} as a coarse geometric layout cue and partition it into overlapping point cloud blocks. Starting from a seed block, we sequentially generate the remaining blocks across three generation stages, conditioning each block on its overlaps with already-generated neighbours (\Cref{sec:Method_World}). The overall pipeline is shown in~\Cref{fig:pipeline}.
In this way, the point-cloud scaffold pins the scene to the observed geometry, while the block-wise generative prior fills in complete, high-fidelity surfaces.

\subsection{Data Curation}
\label{sec:Method_Data}

\noindent\textbf{Scene Slicing.}
To enable block-level scene generation, we first construct a structured, scalable 3D block dataset. We build on two synthetic indoor scene datasets, 3D-FRONT~\cite{3DFront} and SAGE-10k~\cite{Sage}, which together cover diverse layouts, furniture styles, and material distributions. 
For each scene, we render RGBD views from randomised eye-level cameras under varied lighting.
We then propose a scene-slicing strategy. The scene is partitioned into a regular grid of canonically oriented cubic blocks with side length $\ell$, each covering the full vertical extent of the scene. This yields roughly $56\text{k}$ blocks from $6\text{k}$ 3D-FRONT scenes and $61\text{k}$ blocks from $10\text{k}$ SAGE-10k scenes, totalling around $117\text{k}$ training blocks. Each mesh block is normalised to canonical $[-0.5, 0.5]^3$ space. 

\noindent\textbf{Condition Extraction.}
Each block then yields a training tuple $(\mathcal{M}_b,\ \mathcal{P}_b,\ \{(I_v, K_v, R_v, t_v)\}_{v=1}^{16})$, denoting the block-local mesh, point cloud, RGBA renders, and camera poses, respectively.
Every rendered RGBD view is back-projected into the block's canonical space, yielding a block-local camera pose $(K_v, R_v, t_v)$. The pixels falling inside the block cube form the RGBA render $I_v$, with the in-cube mask as alpha. We retain the top-$16$ views per block by in-cube pixel coverage.
For $\mathcal{P}_b$, we randomly sample from two complementary noisy sources during training: unprojected GT depth from a random subset of views with augmented noise, and a VGGT-$\Omega$~\cite{VGGT-Omega} reconstruction from $\{I_v\}_{v=1}^{16}$, both downsampled to $M$ points. This exposes the model to both controllable augmentation and the target inference distribution, keeping generation robust when only VGGT-$\Omega$ point clouds are available at test time.

\subsection{Reconstruction-based Block Generation}
\label{sec:Method_Block}

\noindent\textbf{Generative Prior.}
We build on TRELLIS.2~\cite{TRELLIS2}, an image-to-3D generation model that represents 3D asset as a sparse voxel structure termed \emph{O-Voxel}, defined as a set of feature tuples $\mathbf{f} = \{(\mathbf{f}^{\text{geo}}_i,\, \mathbf{f}^{\text{mat}}_i,\, \mathbf{o}_i)\}_{i=1}^{L}$ over $L$ active voxels on a regular $G^3$ grid, where $\mathbf{o}_i$ denotes the voxel position, $\mathbf{f}^{\text{geo}}_i$ encodes local geometric information, and $\mathbf{f}^{\text{mat}}_i$ encodes material properties. 
Generation cascades three DiT flow-matching stages: \emph{Sparse Structure} predicts the active-voxel positions $\{\mathbf{o}_i\}$. \emph{Geometry} predicts geometry latents $\mathbf{z}^{\text{geo}}$ within active voxels, which are decoded into $\{\mathbf{f}^{\text{geo}}_i\}$. \emph{Material} synthesizes material latents $\mathbf{z}^{\text{mat}}$ aligned to $\mathbf{z}^{\text{geo}}$, decoded to $\{\mathbf{f}^{\text{mat}}_i\}$.
We finetune all three stages from the pretrained object-level TRELLIS.2 for block-level scene generation, replacing its original single-image conditioning with our reconstruction-based conditioning as detailed below.

\noindent\textbf{Reconstruction-Based Conditioning.}
We leverage the block-level conditions curated in \Cref{sec:Method_Data} for controllable 3D generation. Multi-view images and their
camera poses are jointly encoded into a unified representation $\mathbf{c}^{\text{mv}}$ shared by all three stages. \emph{Sparse Structure} stage additionally receives point-cloud tokens $\mathbf{c}^{\text{pc}}$ as a coarse geometric prior, giving $\mathbf{c} = [\mathbf{c}^{\text{mv}}, \mathbf{c}^{\text{pc}}]$. The resulting conditioning is injected into every cross-attention layer of the DiT.

\emph{Multi-View Image Condition.}
For each block, we randomly sample $V$ views $\{I_v\}_{v=1}^{V}$ with their camera poses $\{(R_v, t_v, K_v)\}_{v=1}^{V}$. Each view is encoded by a frozen DINOv3~\cite{DINOv3} patch encoder into image tokens $\mathbf{c}_v^{\text{img}}$, while its camera is represented by five Pl\"ucker rays which are projected through an MLP into camera tokens $\mathbf{c}_v^{\text{cam}}$.
Per-view camera tokens are prepended to the corresponding image patches, forming the multi-view context $\mathcal{S} = [\mathbf{c}_1^{\text{cam}}, \mathbf{c}_1^{\text{img}},\ldots,\mathbf{c}_V^{\text{cam}}, \mathbf{c}_V^{\text{img}}]$. Starting from a set of learnable query tokens $\mathbf{q}_0$, we run $N_t$ transformer cross-attention blocks that progressively fuse $\mathcal{S}$ into the fixed-length tokens:
\begin{equation}
    \label{eq:mv_fusion}
    \mathbf{q}_{l+1} = \text{CrossAttn}\big(Q(\mathbf{q}_l),\ K(\mathcal{S}),\ V(\mathcal{S})\big),\quad l \in \{0, 1, \ldots, N_t-1\},
\end{equation}
yielding the multi-view condition $\mathbf{c}^{\text{mv}} = \mathbf{q}_{N_t}$.

\emph{Point Cloud Condition.}
Recovering geometry from RGB alone is inherently ambiguous, especially at scene scale where sparse views cover only partial structure. We therefore condition the Sparse Structure stage on a point cloud that supplies coarse spatial structure.
Following the Hunyuan3D-Omni~\cite{Hunyuan3D-Omni} design, each point of $\mathcal{P}_b = \{\mathbf{p}_j\}_{j=1}^{M}$ is encoded by a Fourier positional embedder followed by a linear projection, and $K$ learnable modality tokens are appended to form the point-cloud condition $\mathbf{c}^{\text{pc}}$.
Point-cloud tokens are randomly dropped during training, allowing classifier-free guidance to support both point-free and point-conditioned inference.

\noindent\textbf{Neighbour-Aware Latent Inpainting.}
At inference, the scene is extended block-by-block. Each new block only observes partial 3D context from its already-generated neighbours, yet it must produce a complete 3D representation that continues the surrounding geometry and appearance seamlessly. We formulate this as a 3D block inpainting task and train dedicated inpaint variants of all three denoisers. We augment each denoiser's native input with $C{+}1$ inpainting-specific channels, comprising a one-channel binary inpainting mask $\mathbf{m}$ and a $C$-channel masked known latent $\mathbf{x}_0\odot(\mathbf{1}-\mathbf{m})$, where $C$ denotes the latent feature dimension. The resulting hidden feature is:
\begin{equation}
  \label{eq:inpaint_input}
  \mathbf{h}_\text{in}
  =
  f_\text{in}\Big(
  [\,\mathbf{x}_t\ [,\mathbf{z}^{\text{geo}}],\ \mathbf{m},\
  \mathbf{x}_0\odot(\mathbf{1}-\mathbf{m})\,]
  \Big).
\end{equation}
Here, $\mathbf{m}_i=0$ marks known latent tokens supplied, whereas $\mathbf{m}_i=1$ marks tokens to be freshly generated. $\mathbf{z}^{\text{geo}}$ denotes an optional \textit{Geometry} latent included for the \textit{Material} stage. 

To retain the priors of our block-level generation models, we initialise each inpaint variant from its corresponding generation model. 
The original input projection is directly inherited, while the projection weights associated with the additional $C{+}1$ inpainting channels are zero-initialised. At the start of finetuning, the inpaint variant is thus functionally equivalent to the base generator, providing a stable initialisation for optimisation.

Training masks are sampled to emulate the overlap layouts arising at inference. Since all scenes fit within a single block along the vertical axis, we restrict masking to the $XY$ plane. On each iteration, we sample the number of neighbour-supplied edges $E \in \{1, 2\}$ of the block, enforcing an \emph{L-shape} constraint (one $X$-edge plus one $Y$-edge) when $E=2$. The mask configurations cover the $1$- and $2$-neighbour cases encountered under our voxel-priority generation ordering (\Cref{sec:Method_World}).
The known margin along each edge is sampled between $3\ell/16$ and $\ell/4$.   
At inference, we shrink the raw overlap of width $\ell/4$ by one quarter, leaving a known margin of $3\ell/16$ and an editable transition band of width $\ell/16$. This design preserves sufficient neighbour context for cross-block consistency while allowing the model to adjust geometric and material transitions near block boundaries.

\subsection{World Generation Pipeline}
\label{sec:Method_World}
Given an arbitrary number of uncalibrated RGB images $\{I_n\}_{n=1}^{N}$, our inference pipeline generates a large-scale 3D mesh through block-based, neighbour-aware latent inpainting and overlap blending.

\noindent\textbf{Block Preparation.}
We run VGGT-$\Omega$~\cite{VGGT-Omega} to recover coarse camera parameters and a global point cloud from the input RGB images. After confidence filtering and outlier removal, we align and rescale the reconstruction to the canonical scene coordinates.
We partition its $XY$ bounding box into overlapping cubic blocks of side length $\ell$, with adjacent blocks overlapping by $\ell/4$. 
For each block, we retain only the in-block pixels of each input RGB image and consider images with at least $15\%$ in-block pixel coverage as candidate views. We voxelise the corresponding visible points and select up to $V$ views whose union maximises the block's voxel coverage. This produces the training-compatible tuple  $(\hat{\mathcal{P}}_b, \{(I_v,\hat{K}_v,\hat{R}_v,\hat{t}_v)\}_{v=1}^{V})$, without requiring ground-truth geometry or calibrated cameras.

We determine a voxel-priority generation order based on block-wise voxel occupancy. Starting from the block with the highest occupancy, we greedily expand a connected frontier while ensuring that each block has at most two previously generated four-connected neighbours. This ordering matches the $1$- and $2$-neighbour masks used to train the block inpainters.


\noindent\textbf{Sparse Structure World Generation.}
We maintain shared world-space \emph{Sparse Structure} latent and occupancy. 
The seed latent is sampled by the generator, while subsequent blocks are inpainted from overlaps with generated neighbours.
To ensure \emph{Sparse Structure} continuity across block boundaries, we introduce a \textit{trajectory-clamping} mechanism. At every denoising step $t$, the known overlap is restored to its forward-flow trajectory:
\begin{equation}
    \label{eq:ss_known_trajectory}
    \mathbf{x}^{\mathrm{ss}}_{t,\mathrm{known}}
    =
    (1-t)\mathbf{x}^{\mathrm{ss}}_{0,\mathrm{known}}
    +
    t\boldsymbol{\epsilon}.
\end{equation}
Here, $\mathbf{x}^{\mathrm{ss}}_{0,\mathrm{known}}$ is the clean latent extracted from the world-space latent. We then clamp the flow state $\mathbf{x}^{\mathrm{ss}}_t$ predicted by the inpainter by replacing its known overlap:
\begin{equation}
    \label{eq:ss_overlap_clamp}
    \tilde{\mathbf{x}}^{\mathrm{ss}}_{t}
    =
    \mathbf{m}^{\mathrm{ss}}\odot\mathbf{x}^{\mathrm{ss}}_{t}
    +
    (\mathbf{1}-\mathbf{m}^{\mathrm{ss}})\odot
    \mathbf{x}^{\mathrm{ss}}_{t,\mathrm{known}}.
\end{equation}

This mechanism preserves the previously generated overlap while allowing the unknown region to evolve along the denoising trajectory. After denoising, the resulting clean latent $\mathbf{z}^{\mathrm{ss}}=\tilde{\mathbf{x}}^{\mathrm{ss}}_0$ is decoded into an occupancy grid $\mathbf{o}^{\mathrm{ss}} = \mathcal{D}_{\mathrm{ss}}(\mathbf{z}^{\mathrm{ss}})$. Both are written only to unoccupied locations in the shared world-space tensors, which then provide context for subsequent blocks.

\noindent\textbf{Geometry and Material World Generation.}
Given the completed world \textit{Sparse Structure} latent, we process blocks in the same causal order. For each block, we generate or inpaint its \textit{Geometry} latent, followed by its \textit{Material} latent. 
To reduce block-boundary seams, each block's clean latent $\mathbf{z}^{s}_{b}$, where $s\in\{\mathrm{geo},\mathrm{mat}\}$, is smoothly blended into the corresponding world latent $\mathbf{Z}^{s}_{b}$ immediately after generation. 
The blending weights are determined by the distance to the block boundaries.
Since blocks overlap only in the $XY$ plane, let $\mathcal{A}\subseteq\{x,y\}$ denote the overlapping axes, $\mathbf{u}=(x,y,z)$ the world-grid coordinate, and $\boldsymbol{\xi}_b(\mathbf{u})\in\{0,\ldots,G-1\}^3$ the corresponding block-local latent-voxel coordinate, where $G$ is the latent-grid resolution along each axis. As adjacent blocks overlap by one quarter of the block, the overlap spans $G/4$ latent voxels. We define the linear feather weight as
\begin{equation}
    \label{eq:world_feather_weight}
    \alpha_b(\mathbf{u})
    =
    \prod_{a\in\mathcal{A}}
    \operatorname{clip}\!\left(
    \frac{
    \min\{\xi_{b,a}(\mathbf{u})+\tfrac{1}{2},\ G-\xi_{b,a}(\mathbf{u})-\tfrac{1}{2}\}
    }{G/4},
    0,1
    \right).
\end{equation}
This produces smooth linear transitions across seams and bilinear blending at block corners. Let $\mathbf{Z}^{s}_{b}(\mathbf{u})$ denote the world latent after integrating block $b$, and let $A_b(\mathbf{u})$ denote the accumulated weight, with $A_0(\mathbf{u})=0$. For each active world-grid location $\mathbf{u}$, we update them online as
\begin{equation}
    \label{eq:world_weighted_writeback}
    \mathbf{Z}^{s}_{b}(\mathbf{u})
    =
    \frac{
    A_{b-1}(\mathbf{u})\mathbf{Z}^{s}_{b-1}(\mathbf{u})
    +
    \alpha_b(\mathbf{u})\mathbf{z}^{s}_b(\mathbf{u})
    }{
    A_{b-1}(\mathbf{u})+\alpha_b(\mathbf{u})
    },
    \qquad
    A_{b}(\mathbf{u})
    =
    A_{b-1}(\mathbf{u})+\alpha_b(\mathbf{u}).
\end{equation}
Once the world latents are complete, we extract the aligned \textit{Geometry} and \textit{Material} latents for each block and decode them into block-local meshes with PBR materials. The block meshes are then transformed into the shared world frame and merged sequentially.
\section{Experiments}
\label{sec:Experiments}

\subsection{Experimental Setup}
\label{sec:Exp_Setup}

\noindent\textbf{Datasets.}
We evaluate on unseen scenes from two datasets. From 3D-FRONT~\cite{3DFront}, we hold out 30 scenes to assess performance on synthetic data. From ScanNet++~\cite{Scannet++}, we evaluate 30 scenes to assess out-of-domain real-world performance. Each scene provides 32 unposed RGB images as input and 50 held-out RGB, depth, and normal views for evaluation.

\noindent\textbf{Metrics.}
We evaluate reconstructed meshes in both 3D and 2D. In 3D, we measure geometric accuracy and coverage using Chamfer distance (CD), F-score (\(10\) cm), and normal consistency (NC, thresholded at \(20\) cm). In 2D, we report geometric errors (RMSE, AbsRel, and angular normal error), computed over mutually valid pixels; perceptual and semantic metrics (LPIPS and CLIP) on depth and normal renderings; and completeness over valid reference pixels.

\noindent\textbf{Baselines.}
We compare against five representative baselines spanning different reconstruction paradigms. 
\textit{\textbf{VGGT-$\Omega$}}~\cite{VGGT-Omega} predicts per-view depth maps and camera poses, then fuses them into a mesh via TSDF integration~\cite{TSDF} and Marching Cubes. All of the following methods consume its predictions as input. 
\textit{\textbf{Murre}}~\cite{Murre} conditions a large monocular depth estimator on the predicted point cloud and fuses the resulting depth into a mesh.
\textit{\textbf{MonoSDF}}~\cite{MonoSDF} optimizes a per-scene signed distance field guided by monocular depth and normal priors.
\textit{\textbf{FineRecon}}~\cite{FineRecon} performs feed-forward TSDF regression conditioned on multi-view depth guidance. 
\textit{\textbf{GenRecon}}~\cite{GenRecon} tiles the scene into overlapping chunks and projects multi-view image features into per-chunk 3D feature volumes.

\noindent\textbf{Implementation details.}
We partition scenes into \(4\,\mathrm{m}\) cubic blocks, each conditioned on up to \(4\) local views and an \(8192\)-point cloud crop. Starting from the \(512\)-resolution TRELLIS.2 pretrain model, we fully finetune the \textit{Sparse Structure} and
\textit{Geometry} stages and apply LoRA~\cite{LoRA} to \textit{Material} stage. At inference, we use a \(12\)-step
Heun sampler with guidance scales of \(7.5\), \(7.5\), and \(3.0\) and time-rescaling factors of \(5.0\), \(3.0\), and \(3.0\). All models are trained on \(32\) NVIDIA GPUs. Details
are provided in Appendix.

\subsection{Quantitative Results}
\label{sec:Exp_Quantitative}

\begin{table*}[t]
  \scriptsize
  \caption{\textbf{Comparisons on Synthetic Scenes.} We evaluate 3D geometry and 2D depth and normal metrics on \(30\) scenes from 3D-FRONT, which is synthetic and large-scale. }
  \vspace{-7pt}
  \label{tab:3dfront_results}
  \centering
  \setlength{\tabcolsep}{3pt}
  \begin{tabular}{@{}lccccccccccc@{}}
    \toprule
    \multirow{2}{*}{\textbf{Method}} &
    \multicolumn{3}{c}{\textbf{3D Geometry}} &
    \multicolumn{1}{c}{\textbf{Coverage}} &
    \multicolumn{4}{c}{\textbf{Depth}} &
    \multicolumn{3}{c}{\textbf{Normal}} \\
    \cmidrule(lr){2-4}\cmidrule(lr){5-5}
    \cmidrule(lr){6-9}\cmidrule(lr){10-12}
    &
    \textbf{CD\(\downarrow\)} &
    \textbf{F-score\(\uparrow\)} &
    \textbf{NC\(\uparrow\)} &
    \textbf{Complete.\(\uparrow\)} &
    \textbf{RMSE\(\downarrow\)} &
    \textbf{AbsRel\(\downarrow\)} &
    \textbf{LPIPS\(\downarrow\)} &
    \textbf{CLIP\(\uparrow\)} &
    \textbf{Angular\(\downarrow\)} &
    \textbf{LPIPS\(\downarrow\)} &
    \textbf{CLIP\(\uparrow\)} \\
    \midrule
    VGGT-\(\Omega\)+TSDF & 0.2934 & 0.4596 & 0.5433 & 0.7363 & 0.5906 & 0.1548 & 0.4873 & 0.8679 & 20.9899 & 0.6188 & 0.8568 \\
    Murre & 0.5295 & 0.2426 & 0.4729 & 0.4572 & 0.8847 & 0.2397 & 0.5273 & 0.8332 & 22.6896 & 0.6671 & 0.8542 \\
    MonoSDF & 0.2558 & 0.4936 & 0.5730 & 0.9056 & 0.6052 & 0.1616 & 0.2566 & 0.9161 & 18.2223 & 0.3560 & 0.9062 \\
    FineRecon & 0.2448 & 0.5301 & \underline{0.6252} & 0.8126 & 0.5578 & \underline{0.1490} & 0.3398 & 0.8906 & \underline{17.6099} & 0.4419 & 0.8845 \\
    GenRecon & \underline{0.1861} & \underline{0.6196} & 0.6128 & \underline{0.9587} & \underline{0.5348} & 0.1666 & \underline{0.2548} & \underline{0.9490} & 17.9822 & \underline{0.3139} & \underline{0.9543} \\
    \rowcolor{lightblue}
    \textbf{\MethodName{} (Ours)} &
    \textbf{0.1583} & \textbf{0.7271} & \textbf{0.6839} &
    \textbf{0.9589} & \textbf{0.3910} & \textbf{0.1126} &
    \textbf{0.1966} & \textbf{0.9582} & \textbf{16.4044} &
    \textbf{0.2951} & \textbf{0.9601} \\
    \bottomrule
  \end{tabular}
\end{table*}

\noindent\textbf{Synthetic scene reconstruction.}
We evaluate 2D and 3D reconstruction performance on 3D-FRONT~\cite{3DFront} using \(32\) unposed RGB images. The evaluation set primarily comprises large-scale multi-room scenes, with an average of \(7.1\) rooms and \(62.1\, \mathrm{m}^2\) of floor area. \MethodName{} achieves the best overall performance, completing regions missing from the initial VGGT-\(\Omega\) reconstruction. It substantially outperforms all baselines across the 3D geometry metrics while achieving the best results on all evaluated 2D depth and normal metrics, demonstrating accurate, consistent, and complete scene reconstruction.
\begin{table*}[t]
  \scriptsize
  \caption{\textbf{Comparisons on Real-world Scenes.}
  We evaluate 3D geometry and 2D depth and normal metrics on \(30\) scenes from Scannet++. Best results are \textbf{bold}, and second-best results are \underline{underlined}.}
  \vspace{-7pt}
  \label{tab:scannet_results}
  \centering
  \setlength{\tabcolsep}{3pt}
  \begin{tabular}{@{}lccccccccccc@{}}
    \toprule
    \multirow{2}{*}{\textbf{Method}} &
    \multicolumn{3}{c}{\textbf{3D Geometry}} &
    \multicolumn{1}{c}{\textbf{Coverage}} &
    \multicolumn{4}{c}{\textbf{Depth}} &
    \multicolumn{3}{c}{\textbf{Normal}} \\
    \cmidrule(lr){2-4}\cmidrule(lr){5-5}
    \cmidrule(lr){6-9}\cmidrule(lr){10-12}
    &
    \textbf{CD\(\downarrow\)} &
    \textbf{F-score\(\uparrow\)} &
    \textbf{NC\(\uparrow\)} &
    \textbf{Complete.\(\uparrow\)} &
    \textbf{RMSE\(\downarrow\)} &
    \textbf{AbsRel\(\downarrow\)} &
    \textbf{LPIPS\(\downarrow\)} &
    \textbf{CLIP\(\uparrow\)} &
    \textbf{Angular\(\downarrow\)} &
    \textbf{LPIPS\(\downarrow\)} &
    \textbf{CLIP\(\uparrow\)} \\
    \midrule
    VGGT-\(\Omega\)+TSDF & \underline{0.1121} & \underline{0.7467} & 0.7036 & 0.8737 & \underline{0.3984} & \underline{0.1433} & 0.3907 & 0.8991 & 25.4852 & 0.5075 & 0.8888 \\
    Murre & 0.1756 & 0.5640 & 0.6222 & 0.5567 & 0.6619 & 0.1863 & 0.5587 & 0.8690 & 28.4193 & 0.5642 & 0.8761 \\
    MonoSDF & 0.1665 & 0.6224 & 0.6913 & 0.9670 & 0.6293 & 0.3043 & \underline{0.2705} & 0.9273 & 23.7527 & \underline{0.3861} & 0.9016 \\
    FineRecon & 0.1259 & 0.6861 & \underline{0.7226} & 0.9388 & 0.4405 & 0.1673 & 0.2707 & 0.9183 & \underline{22.6830} & 0.4043 & 0.8886 \\
    GenRecon & 0.1227 & 0.5901 & 0.5969 & \underline{0.9722} & 0.6837 & 0.3048 & 0.3862 & \underline{0.9291} & 31.5722 & 0.4711 & \underline{0.9172} \\
    \rowcolor{lightblue}
    \textbf{\MethodName{} (Ours)} &
    \textbf{0.0638} & \textbf{0.8363} & \textbf{0.7840} &
    \textbf{0.9918} & \textbf{0.3331} & \textbf{0.0992} &
    \textbf{0.2139} & \textbf{0.9522} & \textbf{20.0429} &
    \textbf{0.3631} & \textbf{0.9320} \\
    \bottomrule
  \end{tabular}
\end{table*}

\noindent\textbf{Real-world scene reconstruction.} We further evaluate generalization on ScanNet++~\cite{Scannet++}. Despite being trained solely on synthetic scenes, \MethodName{} reduces CD by \(43.1\%\) and improves F-score by \(12.0\%\) over the strongest baselines. Its rendered depth and normal maps also exhibit better geometric accuracy, perceptual similarity, and semantic alignment with the laser-scanned ground truth, demonstrating strong sim-to-real generalization.

\subsection{Qualitative Results}
\label{sec:Exp_Qualitative}

\begin{figure*}[t]
    \centering
      \includegraphics[width=\textwidth]{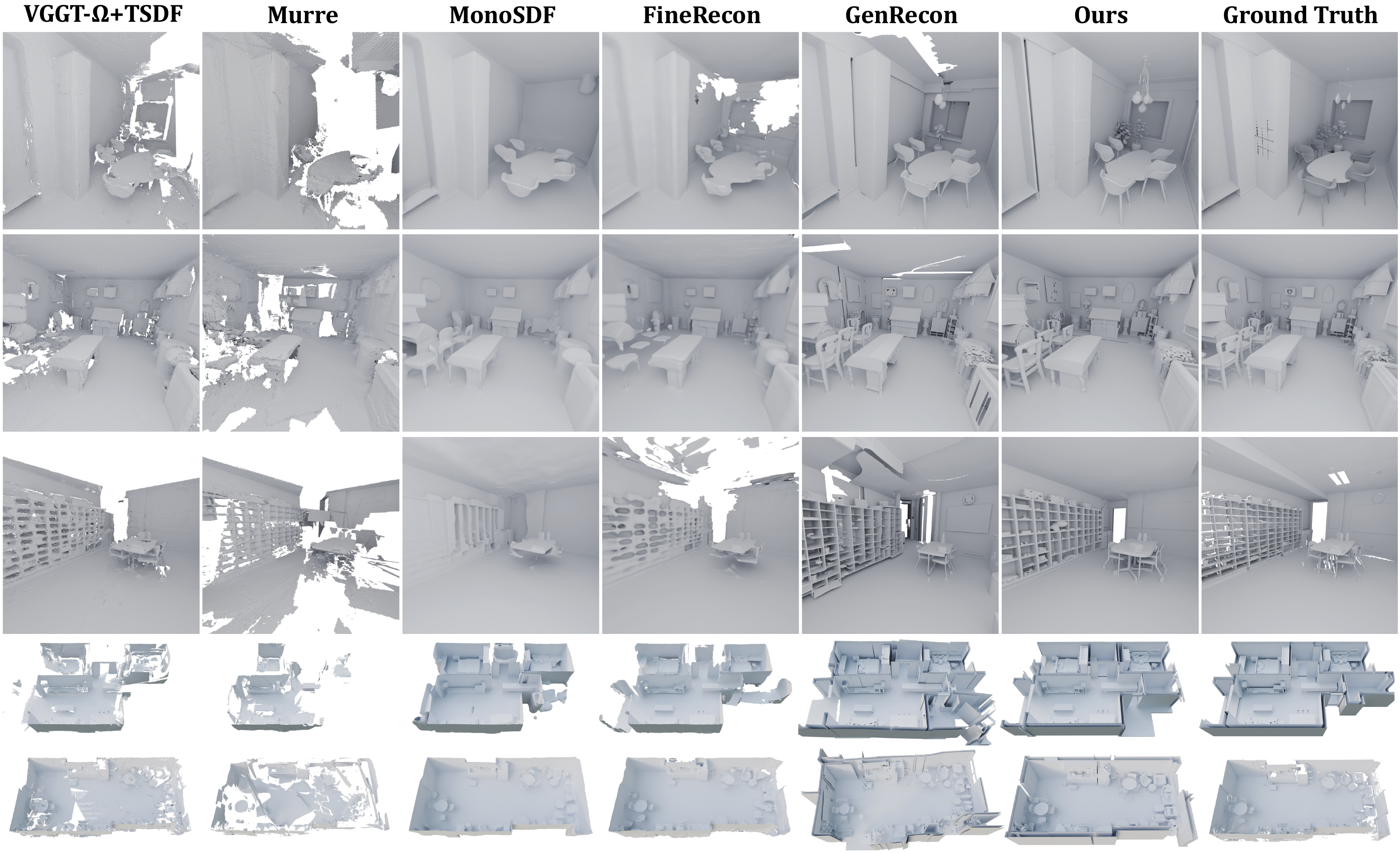}
      \vspace{-15pt}
      \caption{\textbf{Qualitative Comparison.} Mesh reconstruction results on evaluation scenes, including both synthetic and real-world examples. We show close-up views of interior details (top) and complete scene reconstructions (bottom). 
      Compared with the baselines, \MethodName{} produces more complete and geometrically consistent reconstructions with substantially fewer artifacts.
      }
    \label{fig:comparison}
\end{figure*}

We present qualitative comparisons on both synthetic and real-world scenes in~\Cref{fig:comparison}. As a reconstruction-based approach, VGGT-\(\Omega\)+TSDF recovers visible surfaces but leaves unobserved regions incomplete. Deviations in its predicted depths and camera poses can further propagate through TSDF fusion, resulting in noisy artifacts and distorted walls. 
Methods that use geometric predictions from VGGT-\(\Omega\), such as Murre and FineRecon, also exhibit artifacts and broken surfaces.
GenRecon completes some unobserved regions; however, its feature-projection conditioning mechanism injects the distorted point cloud and inaccurate camera poses into the generative model, resulting in fragmented artifacts and distortions of the global scene layout. 
The optimisation-based MonoSDF completes missing regions and reduces geometric noise, but its over-smoothed surfaces lose fine geometric details.
In comparison, \MethodName{} produces more complete and geometrically coherent reconstructions, preserving sharper object boundaries, thin structures, and local details despite noise in the initial geometry and camera estimates.

Beyond geometric reconstruction, \MethodName{} produces scene-level meshes with high-fidelity PBR materials. 
As illustrated in~\Cref{fig:teaser}, our method consistently reconstructs detailed geometry and texture from diverse input sources, including real-world captures, synthetic images, and frames from generated videos~\cite{Seedance2}, demonstrating robust generalization across input domains. 
Additional qualitative results are provided in the Appendix. In particular, \Cref{fig:agent} compares \MethodName {} with an agentic baseline.

\subsection{Ablation Study}
\label{sec:Exp_Ablation}

\begin{figure*}[t]
    \centering
    \includegraphics[width=\textwidth]{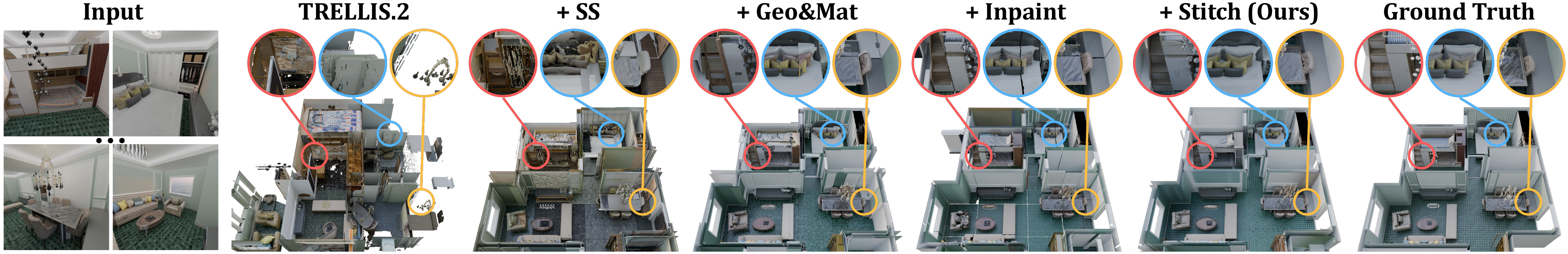}
      \vspace{-15pt}
    \caption{\textbf{Qualitative Ablation Study.} We qualitatively compare the same ablation settings as in~\Cref{tab:ablation} on a 3D-FRONT scene unseen during training. We show the complete scene reconstruction for each variant and the ground truth, while colored circles highlight corresponding local details.}
    \label{fig:ablation}
\end{figure*}

We ablate the key components of \MethodName{}. The ablation experiments are conducted on complete scenes generated from 15 3D-FRONT scenes and 15 ScanNet++ scenes. Starting from the vanilla pretrained TRELLIS.2~\cite{TRELLIS2}, we progressively add four components: reconstruction-based sparse-structure generation (\textit{SS}), reconstruction-based geometry and material generation (\textit{Geo+Mat}), neighbour-aware latent inpainting (\textit{Inpaint}), and stage-wise world latent clamping and blending (\textit{Stitch}). Quantitative results are reported in~\Cref{tab:ablation}, while~\Cref{fig:ablation} visualizes the corresponding effects on representative scene structures and local details.

Although vanilla pretrained TRELLIS.2 has been trained on 3D-FRONT scene data, it performs poorly in our task because its single-image conditioning provides only partial observations and does not enforce a consistent world orientation. Our reconstruction-based \textit{SS} generator instead uses the reconstructed point cloud as a spatial prior while leveraging the generative prior to plausibly complete regions missing from the initial reconstruction. Our reconstruction-based \textit{Geo\&Mat} generators further exploit multi-view observations to recover finer geometry and more faithful materials. Nevertheless, because the blocks are generated independently, inconsistencies can remain in their overlapping regions. As shown in~\Cref{fig:ablation}, adjacent blocks may exhibit misaligned surface heights at their interfaces, accompanied by visible discontinuities in geometry and texture. Adding \textit{Inpaint} propagates context between neighbouring blocks, encouraging their surfaces to connect coherently and improving the depth and normal metrics. However, small inconsistencies at block boundaries may remain visible as seams, explaining the slight degradation in the 3D metrics reported in~\Cref{tab:ablation}.
The complete \MethodName{} pipeline employs stage-wise world latent clamping and blending (\textit{Stitch}) to maintain cross-block continuity and global scene coherence, thereby eliminating visible boundary discontinuities. It consequently achieves the best performance across all reported metrics and produces the reconstruction most consistent with the ground truth.

\begin{table*}[t]
  \scriptsize
  \caption{\textbf{Ablation Study.} We conduct quantitative ablation studies of the key components of \MethodName{} on 30 scenes from 3D-FRONT and ScanNet++. }
  \vspace{-10pt}
  \label{tab:ablation}
  \centering
  \setlength{\tabcolsep}{4.8pt}
  \begin{tabular}{@{}cccccccccccc@{}}
    \toprule
    \multicolumn{4}{c}{\textbf{Components}} &
    \multicolumn{3}{c}{\textbf{3D Metrics}} &
    \multicolumn{5}{c}{\textbf{2D Metrics}} \\
    \cmidrule(lr){1-4}
    \cmidrule(lr){5-7}
    \cmidrule(lr){8-12}
    \textbf{SS} &
    \textbf{Geo\&Mat} &
    \textbf{Inpaint} &
    \textbf{Stitch} &
    \textbf{CD\(\downarrow\)} &
    \textbf{F-score\(\uparrow\)} &
    \textbf{NC\(\uparrow\)} &
    \textbf{Complete.\(\uparrow\)} &
    \textbf{MAE\(\downarrow\)} &
    \textbf{AbsRel\(\downarrow\)} &
    \textbf{Angular\(\downarrow\)} &
    \textbf{RGB CLIP\(\uparrow\)} \\
    \midrule
      \(\times\) & \(\times\) & \(\times\) & \(\times\)
      & 0.4003 & 0.2013 & 0.1682 & 0.7377
      & 1.1847 & 0.6160 & 48.6471 & 0.7255 \\
      \checkmark & \(\times\) & \(\times\) & \(\times\)
      & 0.0996 & 0.7438 & 0.7173 & 0.9722
      & 0.1659 & 0.0980 & 19.6434 & 0.8244 \\
      \checkmark & \checkmark & \(\times\) & \(\times\)
      & \underline{0.0954} & \underline{0.7624} & \underline{0.7349} & 0.9858
      & 0.1589 & 0.0890 & 17.7610 & 0.8609 \\
      \checkmark & \checkmark & \checkmark & \(\times\)
      & 0.1023 & 0.7482 & 0.7202 & \underline{0.9868}
      & \underline{0.1439} & \underline{0.0830}
      & \underline{17.2462} & \underline{0.8609} \\
    \rowcolor{lightblue}
      \checkmark & \checkmark & \checkmark & \checkmark
      & \textbf{0.0830} & \textbf{0.8226} & \textbf{0.7640}
      & \textbf{0.9894}
      & \textbf{0.1394} & \textbf{0.0815}
      & \textbf{16.5789} & \textbf{0.8671} \\
    \bottomrule
  \end{tabular}
\end{table*}

\section{Conclusion}
\label{sec:Conclusion}

We presented \MethodName{}, a generative reconstruction framework for recovering complete, large-scale scene PBR meshes from unposed RGB images. \MethodName{} leverages coarse geometry and camera estimates predicted by a reconstruction model as effective global grounding despite their imperfections. Based on this global scaffold, reconstruction-based block generation produces faithful mesh blocks with detailed geometry and PBR materials. Neighbour-aware latent inpainting and the causal world-generation pipeline stitch these blocks together, ensuring consistent transitions across block boundaries. \MethodName{} achieves state-of-the-art performance in large-scale indoor scene reconstruction, with strong sim-to-real generalization despite training exclusively on synthetic data. More broadly, our work provides a practical path toward reconstructing editable 3D scene assets from casually captured or generated videos, supporting downstream applications such as simulation, embodied AI, and content creation.

\noindent\textbf{Limitations and Future Work.}
As our training data comprise only indoor scenes, the current applicability of \MethodName{} is limited to indoor environments, and its generalization to outdoor scenes remains to be established. In addition, our current scene representation comprises a single layer of \(4\,\mathrm{m}\)-high blocks, restricting reconstruction to single-story environments and precluding the modeling of multi-story buildings. Extending the training distribution to more diverse indoor, outdoor, and multi-story environments, together with supporting block generation along the vertical dimension, constitutes an important direction for future work.

\bibliography{iclr2027_conference}
\bibliographystyle{iclr2027_conference}

\newpage
\section*{Appendix}
\appendix
\setcounter{table}{0}
\setcounter{figure}{0}
\setcounter{equation}{0}
\renewcommand{\thetable}{S\arabic{table}}
\renewcommand{\thefigure}{S\arabic{figure}}
\renewcommand{\theequation}{S\arabic{equation}}
  
The appendix presents supplementary material related to data curation, the
\MethodName{} method, experiments details, and additional
qualitative cases.
\Cref{sec:supp_data} details the block-level training data construction pipeline, including scene rendering, scene slicing, multi-view condition extraction, and point-cloud preparation.
\Cref{sec:method_sup} provides further implementation details of the \MethodName{} framework, covering reconstruction-based conditioning, neighbour-aware latent inpainting, training configurations, and the world generation pipeline.
\Cref{sec:supp_experiments} describes the evaluation protocol, including baseline settings, the alignment procedure, and metric definitions.
\Cref{sec:supp_analysis} presents additional qualitative results, including large-scale scene generation, a comparison with an agentic LLM baseline, and failure case analysis.

\section{Data Curation Details}
\label{sec:supp_data}

We construct the block-level training set from two synthetic indoor scene datasets, 3D-FRONT~\cite{3DFront} and SAGE-10k~\cite{Sage}, as described in the main paper. We detail its rendering, scene-slicing, and condition-extraction procedures below.

\subsection{Scene Rendering}
\label{sec:supp_rendering}

For each scene, we render a set of full-scene RGB observations to simulate the multi-view images used for room reconstruction. We additionally render the corresponding depth maps for point-cloud construction during data preparation.

To increase the appearance diversity of 3D-FRONT, we augment its architectural surfaces with PBR materials from ambientCG (formerly CC0 Textures)\footnote{All assets are released under the CC0 1.0 license: \url{https://ambientcg.com/}.}. For each scene, we construct separate material pools for floors (wood, planks, carpet, tiles, marble, and terrazzo), walls (wallpaper, marble, and plaster), and structural elements such as doors and baseboards (wood and marble). 
We first sample a scene-level material-assignment strategy
\begin{equation}
    s \sim \operatorname{Categorical}(0.3,\,0.6,\,0.1),
    \qquad
    s \in \{\text{\textit{instance}},\,\text{\textit{category}},\,\text{\textit{original}}\}.
\end{equation}
Under the \emph{instance} strategy, each architectural component independently draws a material from its corresponding pool. Under the \emph{category} strategy, components belonging to the same category share a sampled material, producing a more coherent appearance. The \emph{original} strategy applies no additional material augmentation and preserves the scene as provided by 3D-FRONT.
When either \textit{instance} or \textit{category} strategy is selected, floors are always assigned a new material, while walls and structural elements are augmented with probabilities \(0.6\) and \(0.8\), respectively. Furniture is excluded from material augmentation and always retains its original 3D-FUTURE materials.

For illumination, in addition to the original light sources provided by each 3D-FRONT scene, we assign a randomly selected HDRI environment map. SAGE-10k scenes additionally receive one ceiling area light because their enclosed layouts receive limited illumination from the environment map. We randomise the horizontal position, size, energy, and warm/cool colour of each area light.

To mimic natural human capture, we sample near-level camera poses inside individual rooms at heights between \(1.4\text{m}\) and \(1.8\text{m}\). Subject to the availability of valid poses, views are distributed across rooms approximately in proportion to their floor areas. We reject poses closer than \(1\text{m}\) to scene geometry to avoid overly close-up observations and encourage broader scene coverage.

We render RGB and depth maps with BlenderProc~\cite{BlenderProc2} at \(1024\times1024\) resolution and a \(90^\circ\) field of view (FoV). For every accepted view, we also store the camera intrinsics and camera-to-world transformation.

\subsection{Scene Slicing}
\label{sec:supp_slicing}

We divide each scene into overlapping axis-aligned cubic blocks of side length \(\ell=4\text{m}\). For each horizontal axis \(a\in\{x,y\}\), let \([a_{\min},a_{\max}]\) denote the scene extent and \(d_a=a_{\max}-a_{\min}\). The number of blocks \(n_a\), stride \(s_a\), and block starting positions \(p_i^a\) are defined as
\begin{equation}
    \label{eq:supp_scene_slicing}
    n_a=\left\lceil\frac{d_a}{\ell}\right\rceil,\qquad
    s_a=
    \begin{cases}
        0, & n_a=1,\\
        \dfrac{d_a-\ell}{n_a-1}, & n_a>1,
    \end{cases}
    \qquad
    p_i^a=
    \begin{cases}
        \dfrac{a_{\min}+a_{\max}-\ell}{2}, & n_a=1,\\
        a_{\min}+i\,s_a, & n_a>1,
    \end{cases}
\end{equation}
where \(i=0,\ldots,n_a-1\). This construction uses the minimum number of blocks required to cover the scene and distributes their overlap uniformly. We retain only scenes with a height below \(4\text{m}\), so a single block centred on the scene is sufficient along the vertical axis.

For block index \((i,j)\), the starting positions from \Cref{eq:supp_scene_slicing} determine its world-space bounds:
\begin{equation}
    \label{eq:supp_block_bounds}
    \mathbf{b}_{\min}^{i,j}
    =
    \left(
    p_i^x,\,
    p_j^y,\,
    \frac{z_{\min}+z_{\max}-\ell}{2}
    \right),
    \qquad
    \mathbf{b}_{\max}^{i,j}
    =
    \mathbf{b}_{\min}^{i,j}
    +\ell\mathbf{1}.
\end{equation}
We construct each block mesh by clipping the complete scene mesh against four axis-aligned planes of its bounding box. Objects intersecting a block boundary are geometrically cropped rather than assigned wholly to one block, while their original material attributes are preserved. The newly created boundary surfaces are left uncapped.

The block centre is \(\mathbf{c}^{i,j}=(\mathbf{b}_{\min}^{i,j}+\mathbf{b}_{\max}^{i,j})/2\). A world-space point \(\mathbf{x}\) is then mapped to the canonical block frame by
\begin{equation}
    \label{eq:supp_normalize}
    \bar{\mathbf{x}}
    = \frac{\mathbf{x}-\mathbf{c}^{i,j}}{\ell},
\end{equation}
such that every block occupies \([-0.5,0.5]^3\). The same scale and translation are applied to meshes, point clouds, and camera-to-world translations.

\subsection{Multi-View Condition Extraction}
\label{sec:supp_views}

For each rendered image \(I_v\), let \(\mathbf{u}=(u_x,u_y)^\top\), \(\tilde{\mathbf{u}}=(u_x,u_y,1)^\top\), and \(d_v(\mathbf{u})\) denote a pixel, its homogeneous coordinate, and its metric depth, respectively.
Given the camera intrinsics \(\mathbf{K}_v\) and camera-to-world transformation \([\mathbf{R}_v\mid\mathbf{t}_v]\), we unproject the pixel into world space as
\begin{equation}
    \label{eq:supp_unprojection}
    \mathbf{x}_v(\mathbf{u})
    =
    \mathbf{R}_v
    \left(
    d_v(\mathbf{u})\,\mathbf{K}_v^{-1}\tilde{\mathbf{u}}
    \right)
    +\mathbf{t}_v.
\end{equation}
Given the bounds \([\mathbf{b}_{\min}^{i,j},\mathbf{b}_{\max}^{i,j}]\) of block \((i,j)\), we define its visibility mask as
\begin{equation}
    \label{eq:supp_view_mask}
    M_{i,j,v}(\mathbf{u})
    =
    \mathbf{1}\!\left[
    \mathbf{b}_{\min}^{i,j}
    \leq
    \mathbf{x}_v(\mathbf{u})
    \leq
    \mathbf{b}_{\max}^{i,j}
    \right].
\end{equation}
Views are ranked by \(\sum_{\mathbf{u}}M_{i,j,v}(\mathbf{u})\), and only those for which at least \(25\%\) of the image pixels fall inside the block are considered valid. We retain up to \(16\) valid views and use \(M_{i,j,v}\) as the alpha channel of the corresponding RGB image.

We transform the camera-to-world translation to the canonical block frame using \Cref{eq:supp_normalize}, while preserving the camera rotation. Camera intrinsics are rescaled to the resolution used by the image encoder. Thus, the RGBA images, cameras, point cloud, and target latents are expressed in the same block-local coordinate frame.

\subsection{Point-Cloud Preparation}
\label{sec:supp_pointcloud}

For each block, we construct two point-cloud conditions. 
To obtain the augmented depth-derived cloud, we randomly select 6--10 retained views and apply depth and camera perturbations, visibility dropout, and sparse outlier injection before unprojection.
We unproject the perturbed views into the canonical block frame, merge and crop the points to \([-0.5,0.5]^3\), and sparsify the cloud to \(8192\) points.
The VGGT-\(\Omega\) cloud is reconstructed from the retained block-masked RGB views described above. We first apply confidence filtering and statistical outlier removal. We then estimate a Horn Sim(3) transformation from the corresponding VGGT-predicted and canonical camera centres, jointly recovering the relative scale, rotation, and translation between the two coordinate frames. The estimated transformation is applied to the reconstructed point cloud, which is subsequently cropped to the canonical block bounds and sparsified to \(8192\) points.

When both sources are available, they are registered as separate training instances, ensuring balanced exposure to both point-cloud distributions within each epoch. For blocks with only one valid source, the available point cloud is used.

\section{\MethodName{} Method Details}
\label{sec:method_sup}

\subsection{Reconstruction-Based Conditioning}
\label{sec:supp_conditioning}

\noindent\textbf{Multi-View Image Conditioning.} For each conditioning image, we extract patch features using a frozen DINOv3 ViT-L/16 encoder~\cite{DINOv3} at \(512\times512\) resolution. 

To encode the camera pose, we sample five image-plane locations comprising the image centre and four corners. Given a homogeneous pixel coordinate \(\tilde{\mathbf{u}}_{v,k}\), camera intrinsics \(\mathbf{K}_v\), camera-to-world rotation \(\mathbf{R}_v\), and camera centre \(\mathbf{o}_v\), the corresponding world-space ray and Pl\"ucker coordinate are
\begin{equation}
    \mathbf{d}_{v,k}
    =
    \frac{\mathbf{R}_v\mathbf{K}_v^{-1}\tilde{\mathbf{u}}_{v,k}}
    {\left\|\mathbf{R}_v\mathbf{K}_v^{-1}\tilde{\mathbf{u}}_{v,k}\right\|_2},
    \qquad
    \boldsymbol{\pi}_{v,k}
    =
    \left(
    \mathbf{d}_{v,k},
    \mathbf{o}_v\times\mathbf{d}_{v,k}
    \right).
\end{equation}
A two-layer MLP projects each Pl\"ucker coordinate into a camera token, which is prepended to the corresponding image-patch tokens. A four-layer cross-attention module then uses learnable queries to fuse all views into a fixed-length conditioning sequence.

\noindent\textbf{Point Cloud Condition.} The SS stage additionally receives the XYZ coordinates of an \(8192\)-point block cloud \(\mathcal{P}_b=\{\mathbf{p}_j\}_{j=1}^{8192}\). Following Hunyuan3D-Omni~\cite{Hunyuan3D-Omni}, we encode the normalised coordinates using Fourier features and append learned modality tokens:
\begin{equation}
  \label{eq:supp_pc_encoding}
  \mathbf{c}^{\mathrm{pc}}
  =
  \left[
  \psi\!\left(\gamma(\mathbf{p}_1)\right),
  \ldots,
  \psi\!\left(\gamma(\mathbf{p}_{8192})\right),
  \mathbf{e}_1,\ldots,\mathbf{e}_{10}
  \right],
\end{equation} where \(\gamma(\cdot)\) denotes the Fourier positional encoding, \(\psi(\cdot)\) is a learnable projection, and \(\mathbf{e}_{1:10}\) are modality tokens. All tokens share the \(1024\)-dimensional feature space of the image condition. We drop the complete point-cloud stream with probability \(0.1\) during training.

\subsection{Neighbour-Aware Latent Inpainting}
\label{sec:supp_inpainting}

We use the concat-inpainting formulation and overlap masks defined in the main paper for all three stages. Let \(\mathbf{x}^{s}_{0,\mathrm{known}}=\mathbf{x}^{s}_0\odot(\mathbf{1}-\mathbf{m}^{s})\) denote the known latent provided by previously generated neighbours for stage
\(s\in\{\mathrm{ss},\mathrm{geo},\mathrm{mat}\}\), where \(\mathbf{m}^{s}=0\) marks known locations and \(\mathbf{m}^{s}=1\) marks locations to be generated. The input features for the three inpainters are defined separately as
\begin{equation}
  \label{eq:supp_inpaint_ss}
  \mathbf{h}^{\mathrm{ss}}_{\mathrm{in}}
  =
  f^{\mathrm{ss}}_{\mathrm{in}}\!\left(
  \left[
  \mathbf{x}^{\mathrm{ss}}_t,\,
  \mathbf{m}^{\mathrm{ss}},\,
  \mathbf{x}^{\mathrm{ss}}_{0,\mathrm{known}}
  \right]\right),
\end{equation}
\begin{equation}
  \label{eq:supp_inpaint_geo}
  \mathbf{h}^{\mathrm{geo}}_{\mathrm{in}}
  =
  f^{\mathrm{geo}}_{\mathrm{in}}\!\left(
  \left[
  \mathbf{x}^{\mathrm{geo}}_t,\,
  \mathbf{m}^{\mathrm{geo}},\,
  \mathbf{x}^{\mathrm{geo}}_{0,\mathrm{known}}
  \right]\right),
\end{equation}
\begin{equation}
  \label{eq:supp_inpaint_mat}
  \mathbf{h}^{\mathrm{mat}}_{\mathrm{in}}
  =
  f^{\mathrm{mat}}_{\mathrm{in}}\!\left(
  \left[
  \mathbf{x}^{\mathrm{mat}}_t,\,
  \mathbf{z}^{\mathrm{geo}},\,
  \mathbf{m}^{\mathrm{mat}},\,
  \mathbf{x}^{\mathrm{mat}}_{0,\mathrm{known}}
  \right]\right).
\end{equation}
Here, \(\mathbf{z}^{\mathrm{geo}}\) is the native Geometry condition of the
Material stage.

\begin{table*}[t]
\centering
\small
\setlength{\tabcolsep}{4.5pt}
\caption{\textbf{Training configurations}. We report the image and latent
resolutions, point-cloud size, number of GPUs, global batch size, learning
rate, EMA decay, and adaptation strategy for each generation and inpainting
stage.}
\label{tab:supp_training}
\begin{tabular}{lcccccccc}
\hline
Stage & Image & Latent & Points & GPUs & Global Batch & LR & EMA & Adaptation \\
\hline
SS generation & \(512\) & \(16^3\) & \(8192\) & \(32\) & \(128\) & \(1{\times}10^{-4}\) & \(0.9999\) & full \\
SS inpainting & \(512\) & \(16^3\) & \(8192\) & \(32\) & \(128\) & \(3{\times}10^{-5}\) & \(0.9999\) & full \\
Geometry generation & \(512\) & \(32^3\) & -- & \(16\) & \(64\) & \(1{\times}10^{-4}\) & \(0.9995\) & full \\
Geometry inpainting & \(512\) & \(32^3\) & -- & \(16\) & \(64\) & \(3{\times}10^{-5}\) & \(0.9995\) & full \\
Material generation & \(512\) & \(32^3\) & -- & \(16\) & \(96\) & \(3{\times}10^{-5}\) & \(0.9995\) & LoRA \\
Material inpainting & \(512\) & \(32^3\) & -- & \(16\) & \(96\) & \(1{\times}10^{-5}\) & \(0.9995\) & LoRA \\
\hline
\end{tabular}
\end{table*}
\subsection{Training Details}
\label{sec:supp_training}

\Cref{tab:supp_training} summarises the final \(512\)-resolution configurations used for the reported reconstruction results. All models are trained with AdamW and BF16 automatic mixed precision. We use adaptive gradient clipping at the \(95\)-th percentile with a maximum norm of \(1.0\), and unconditional condition dropout of \(0.1\). Geometry and Material training samples are capped at \(16\text{k}\) active sparse tokens.
The Sparse Structure and Geometry stages are fully fine-tuned, whereas the Material denoisers are adapted using rank-\(8\), scale-\(16\) LoRA modules inserted into their attention projections. For the Material stages, the frozen base denoiser is accompanied by trainable multi-view fusion and camera-tokenization modules, whose learning rates are \(10^{-4}\) for generation and \(3{\times}10^{-5}\) for inpainting. The table reports the corresponding LoRA learning rates.

\subsection{World Generation Pipeline}
\label{sec:supp_world}

\noindent\textbf{Global Reconstruction and Scene Alignment.} Given \(N\) unposed RGB images, we resize them to the \(512\)-pixel input resolution of VGGT-\(\Omega\). A single forward pass predicts a depth map \(\hat{\mathbf{D}}_v\), depth confidence \(\hat{\mathbf{C}}_v\), camera intrinsics \(\hat{\mathbf{K}}_v\), and world-to-camera extrinsics \(\hat{\mathbf{E}}_v=[\hat{\mathbf{R}}_v\mid\hat{\mathbf{t}}_v]\) for each view. We unproject the predicted depths using the corresponding camera parameters to obtain world-space point maps while preserving pixel-to-3D correspondences for constructing block-specific image conditions. We merge points satisfying \(\hat{\mathbf{C}}_v(\mathbf{u})\geq2.0\) across all views and apply statistical outlier removal to suppress isolated depth artifacts. Finally, we apply connected-component filtering to the aligned global point cloud and retain its largest component as the scene scaffold.

The VGGT-\(\Omega\) reconstruction is not necessarily aligned with the canonical scene axes. We therefore rotate the predicted point cloud such that the gravity direction is aligned with the positive \(Z\)-axis and the dominant wall directions are aligned with the \(X\)- and \(Y\)-axes, and then recover its physical scale. We first select an initial up direction from the three PCA axes of the global point cloud:
\begin{equation}
    \label{eq:supp_gravity_axis}
    \bar{\mathbf{y}}
    =
    \frac{\sum_v\hat{\mathbf{R}}_v^\top\mathbf{e}_y}
    {\left\|\sum_v\hat{\mathbf{R}}_v^\top\mathbf{e}_y\right\|_2},
    \qquad
    k^\star=\operatorname*{arg\,max}_{k\in\{1,2,3\}}\left|\mathbf{v}_k^\top\bar{\mathbf{y}}\right|,
    \qquad
    \mathbf{g}_0
    =
    -\operatorname{sign}\!\left(
    \mathbf{v}_{k^\star}^\top\bar{\mathbf{y}}
    \right)\mathbf{v}_{k^\star}.
\end{equation}
Here, \(\bar{\mathbf{y}}\) is the average predicted camera \(Y\)-axis expressed in world coordinates, \(\mathbf{v}_k\) is the \(k\)-th PCA axis, and \(\mathbf{g}_0\) is the selected initial up direction. The negative sign accounts for the OpenCV convention in which the camera \(Y\)-axis points downward. After rotating \(\mathbf{g}_0\) to \(\mathbf{e}_z=(0,0,1)^\top\), we refine the up direction using approximately horizontal surfaces in the lower portion of the scene:
\begin{equation}
    \label{eq:supp_gravity_refine}
    \mathcal{H}
    =
    \{i\in\mathcal{P}:|\mathbf{n}_i^\top\mathbf{e}_z|>\tau_h\},
        \qquad
    \mathbf{g}
    =
    \frac{
    \sum_{i\in\mathcal{H}}
    \operatorname{sign}(\mathbf{n}_i^\top\mathbf{e}_z)
    |\mathbf{n}_i^\top\mathbf{e}_z|\mathbf{n}_i
    }{
    \left\|
    \sum_{i\in\mathcal{H}}
    \operatorname{sign}(\mathbf{n}_i^\top\mathbf{e}_z)
    |\mathbf{n}_i^\top\mathbf{e}_z|\mathbf{n}_i
    \right\|_2
    }.
\end{equation}
Here, \(\mathbf{n}_i\) is the local surface normal, \(\mathcal{H}\) retains normals satisfying the horizontal-surface threshold \(\tau_h=0.7\). Their consistently oriented, weighted average gives the refined up direction \(\mathbf{g}\), which is aligned with \(\mathbf{e}_z\). We then estimate the dominant direction of wall-like normals:
  \begin{equation}
      \label{eq:supp_yaw_alignment}
      \begin{aligned}
          \mathcal{W}
          &=
          \left\{
              i:
              \left|\mathbf{n}_i^\top\mathbf{e}_z\right|
              < \tau_w
          \right\},
          \qquad
          \phi_i
          =
          \operatorname{atan2}(n_{i,y},n_{i,x})
          \bmod \frac{\pi}{2},
          \\[2pt]
          \phi^\star
          &=
          \operatorname*{arg\,max}_{\phi\in[0,\pi/2)}
          \sum_{i\in\mathcal{W}}
          \left\|\mathbf{n}_{i,xy}\right\|_2\,
          K_{\sigma}^{\pi/2}(\phi-\phi_i),
          \qquad
          \mathbf{R}_{\mathrm{yaw}}
          =
          \mathbf{R}_z(-\phi^\star).
      \end{aligned}
  \end{equation}
Here, \(\mathcal{W}\) contains wall-like normals selected with \(\tau_w=0.2\), \(\phi_i\) is the horizontal orientation of normal \(i\) folded modulo \(\pi/2\), and \(K_{\sigma}^{\pi/2}\) is a periodic Gaussian kernel used to aggregate their votes. The dominant angle \(\phi^\star\) defines the yaw rotation \(\mathbf{R}_{\mathrm{yaw}}\).

\noindent\textbf{Block Construction and Local Conditioning.}
We rescale the aligned scene to a vertical extent of \(2.9\,\mathrm{m}\) and cover the \(XY\) bounding box of its connected scaffold with the smallest centred grid of \(4\,\mathrm{m}\) blocks. Adjacent blocks use a \(3\,\mathrm{m}\) stride and share the same vertical centre. We discard blocks containing fewer than \(100\) reconstructed points. Each remaining crop is uniformly subsampled to at most \(8N_{\mathrm{pc}}\) points before farthest-point sampling produces the final \(N_{\mathrm{pc}}=8192\)-point condition.

To select the conditioning views, we voxelise each block crop at \(G^3\) resolution with \(G=32\), obtaining a target set \(\mathcal{Q}_b\). For every candidate view \(v\) satisfying the \(15\%\) in-block coverage criterion, we voxelise its aligned in-block point map and retain the intersection \(\mathcal{V}_{b,v}\) with \(\mathcal{Q}_b\). We then choose up to four views that maximise their joint coverage:
\begin{equation}
    \label{eq:supp_max_union_views}
    \mathcal{S}_b^\star
    =
    \operatorname*{arg\,max}_{\substack{\mathcal{S}\subseteq\mathcal{C}_b\\
    |\mathcal{S}|=\min(4,|\mathcal{C}_b|)}}
    \left|
    \bigcup_{v\in\mathcal{S}}\mathcal{V}_{b,v}
    \right|,
\end{equation}
where \(\mathcal{C}_b\) denotes the candidate-view set.


\section{Experimental Details}
\label{sec:supp_experiments}

\subsection{Baselines}
\label{sec:supp_baselines}

We use VGGT-\(\Omega\)~\cite{VGGT-Omega} prediction as the common geometric prior for all methods. It provides camera intrinsics and extrinsics, per-view depth and confidence, and a point cloud obtained by unprojecting the predicted depths. Each baseline follows its official pipeline, with any required ground-truth geometric inputs replaced by the corresponding
predictions from VGGT-\(\Omega\).
\begin{itemize}[leftmargin=*,nosep]
    \item \textbf{VGGT-\(\Omega\)+TSDF} directly integrates the predicted RGB-D
    frames using the VGGT-\(\Omega\) cameras through TSDF
    fusion~\cite{TSDF}.
    \item \textbf{Murre}~\cite{Murre} receives the RGB frames and sparse depth
    samples selected from the VGGT-\(\Omega\) depth maps using confidence,
    spatial subsampling, and multi-view consistency. Its refined dense depths
    are fused with the VGGT-\(\Omega\) intrinsics and aligned poses using TSDF.
    \item \textbf{FineRecon}~\cite{FineRecon} receives the RGB frames,
    confidence-filtered VGGT-\(\Omega\) depths, intrinsics, and aligned poses.
    The predicted depths are used for depth-guided feature back-projection and
    as an additional TSDF feature channel.
    \item \textbf{MonoSDF}~\cite{MonoSDF} uses the VGGT-\(\Omega\) intrinsics
    and aligned poses together with the RGB frames. Following its original
    formulation, its monocular depth and normal supervision is predicted from
    the RGB images by Omnidata, rather than taken from the VGGT-\(\Omega\)
    depth maps.
    \item \textbf{GenRecon}~\cite{GenRecon} uses the aligned VGGT-\(\Omega\)
    point cloud to determine the scene extent and chunk layout, and uses the
    predicted cameras to project multi-view RGB features into each chunk's 3D
    feature volume.
\end{itemize}

\subsection{Evaluation Protocol}
\label{sec:supp_alignment}

As VGGT-\(\Omega\) estimates camera poses from uncalibrated images, its reconstructed coordinate frame has no absolute scale or canonical horizontal orientation. Our scene-alignment procedure recovers gravity and dominant wall directions, but does not resolve the remaining global similarity ambiguity with respect to the ground-truth frame. For evaluation, we therefore estimate a Sim(3) transform from corresponding predicted and reference camera centres using Horn alignment and apply it to the complete predicted mesh. The same transform is used for all 2D and 3D metrics. We convert dataset-specific axis conventions before computing the metrics.

Following the evaluation protocol of GenRecon~\cite{GenRecon}, for synthetic scenes, we restrict prediction triangles to the GT axis-aligned bounding box enlarged by \(20\%\), preventing geometry far outside the scene extent from dominating the evaluation. For ScanNet++, we restrict the evaluation domain to the observed scanner surface, represented by a \(1.1\,\mathrm{cm}\) voxel grid and dilated by \(15\,\mathrm{cm}\).

\subsection{Metrics}
\label{sec:supp_metrics}

\noindent\textbf{3D Metrics.}
We uniformly sample \(200\text{k}\) points from each mesh surface and compute bidirectional nearest-neighbour correspondences. We report:
\begin{itemize}[leftmargin=*,nosep]
    \item \textbf{Chamfer Distance (CD):} the average of the mean
    predicted-to-reference and reference-to-predicted nearest-neighbour
    Euclidean distances, reported in metres;
    \item \textbf{F-score@10cm (F-score):} the harmonic mean of predicted-to-GT
    precision and GT-to-prediction recall, where a point is considered matched
    if its nearest-neighbour distance is below \(0.1\,\mathrm{m}\);
    \item \textbf{Normal Consistency (NC):} the bidirectional mean absolute
    dot product between the normals of nearest-neighbour surface samples.
    Correspondences farther than \(0.2\,\mathrm{m}\) contribute zero.
\end{itemize}

\noindent\textbf{2D Metrics.}
We render the aligned predicted and reference geometry using the same held-out camera poses and intrinsics. For each scene, we evaluate \(50\) views sampled from the held-out test set. The synthetic test views have a resolution of \(512\times512\), whereas the ScanNet++ test views have a resolution of \(640\times480\). Frames used as input to VGGT-\(\Omega\) are excluded from the evaluation views.

We evaluate pixel-wise depth and surface-normal errors, pooling RMSE, AbsRel, and Angular over all valid view--pixel pairs across the evaluation views. Let \(\mathcal{V}_d\) denote the set of view--pixel pairs for which both the predicted and reference renderings have finite, positive depth. The depth metrics are defined as
\begin{equation}
    \mathrm{RMSE}=\sqrt{\frac{1}{|\mathcal{V}_d|}\sum_{p\in\mathcal{V}_d}(\hat d_p-d_p)^2},\qquad \mathrm{AbsRel}=\frac{1}{|\mathcal{V}_d|}\sum_{p\in\mathcal{V}_d}\frac{|\hat d_p-d_p|}{d_p},
\end{equation}
where \(\hat d_p\) and \(d_p\) denote the predicted and reference depths, respectively.
For surface normals, let \(\mathcal{V}_n\) denote the set of view--pixel pairs for which both the predicted and reference foreground normals are valid. Assuming unit-length normals, we report the mean angular error in degrees:
\begin{equation}
    \mathrm{Angular}=\frac{180}{\pi|\mathcal{V}_n|}\sum_{p\in\mathcal{V}_n}\arccos\!\left(\left|\hat{\mathbf n}_p^\top\mathbf n_p\right|\right).
\end{equation}
The absolute dot product makes the metric invariant to normal sign ambiguity.
Completeness measures the fraction of valid reference foreground pixels for which the predicted rendering contains valid geometry. 

For perceptual evaluation, LPIPS and CLIP similarity are computed separately for depth and surface normals. LPIPS, CLIP similarity, and Completeness are computed independently for each view and then averaged across the evaluation views.

\section{Additional Results}
\label{sec:supp_analysis}

\begin{figure*}[t]
    \centering
    \includegraphics[width=\textwidth]{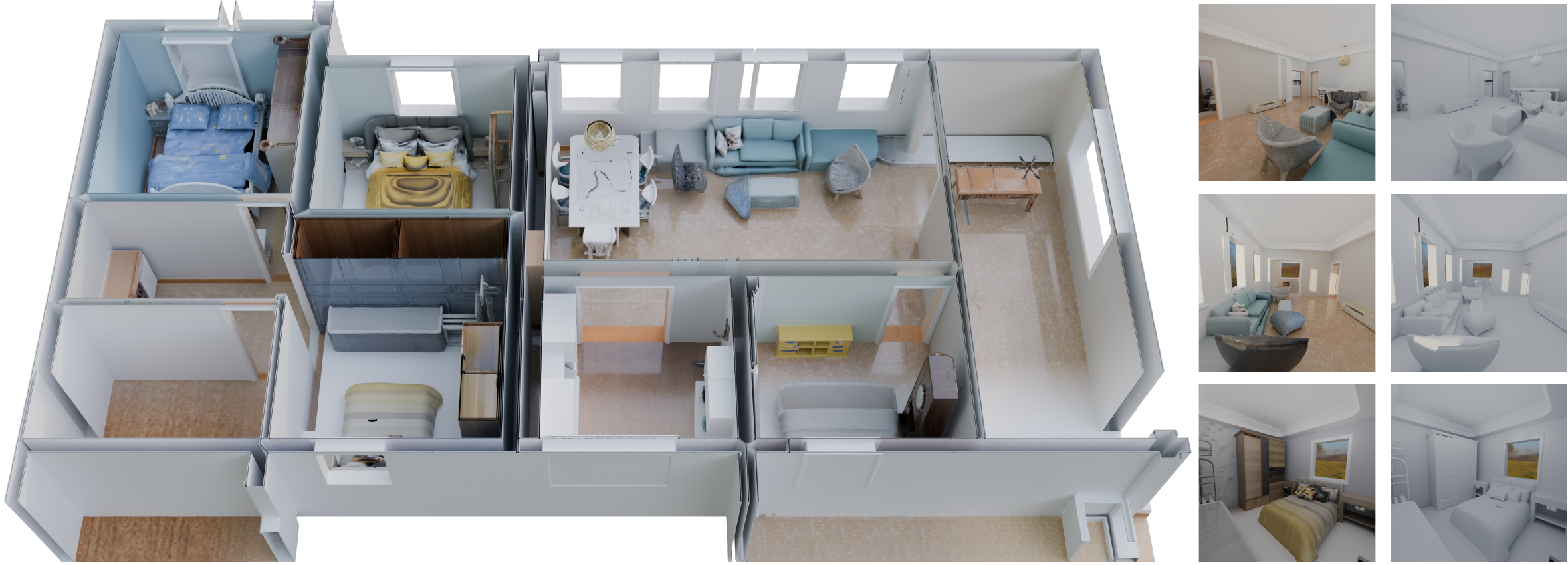}
    \vspace{-15pt}
    \caption{\textbf{Large-Scale Scene Generation.} Reconstruction results for large-scale, multi-room scenes, with top-down views (left) and detailed interior views (right).}
    \label{fig:large_scene}
\end{figure*}

\subsection{Large Scene Generation}
We further demonstrate our method on large-scale indoor scenes comprising multiple, largely isolated rooms. As shown in Fig.~\ref{fig:large_scene}, our method reconstructs an expansive nine-room scene with high fidelity, demonstrating its scalability to large 3D indoor environments.

\begin{figure*}[t]
      \centering
      \includegraphics[width=\textwidth]{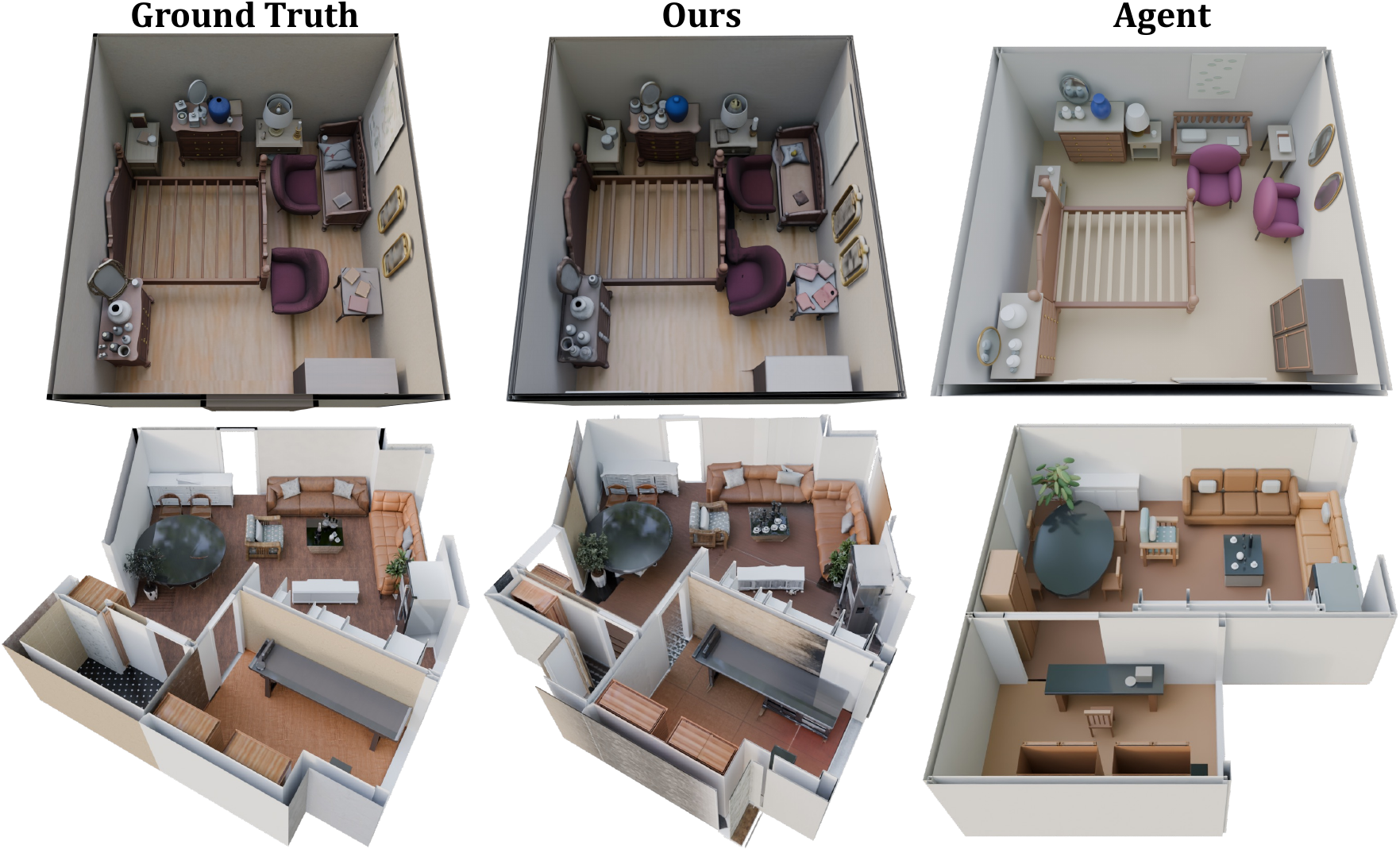}
      \vspace{-15pt}
        \caption{
              \textbf{Comparison with Agentic Baseline.}
              Given the same 32 views, we show GT, ours, and a GPT-6 Astra agent adapted from BVB~\cite{BVB} (left to right). Our method yields more faithful reconstructions, while the agent simplifies or omits objects and distorts the scene layout.
          }
      \label{fig:agent}
  \end{figure*}

\subsection{Comparison with Agentic Baseline.}
We compare our method with a GPT-6 Astra agent configured with high reasoning effort and adapted from BVB's reconstruction protocol~\cite{BVB}. Given the same 32 RGB views, the agent inspects the images, writes and executes Blender scripts, and examines rendered previews to refine its reconstruction. It constructs scenes using geometric primitives and materials specified by numerical parameters, without image textures or imported assets. Our adaptation uses fixed RGB views instead of video input and does not enforce the original cost ceiling.

Figure~\ref{fig:agent} shows a densely furnished scene with complex object arrangements and mutual occlusions (top), and a multi-room interior with oblique partitions (bottom). The recorded reconstruction stages consume approximately 2.12M and 1.71M tokens, respectively, including repeated cached context, corresponding to estimated Standard API-equivalent costs of \$2.93 and \$2.26. In contrast, our method performs inference on a single GPU without external LLM calls, avoiding this additional API expense.
Despite iterative refinement, the agent produces simplified object geometry and misplaced furniture. In the multi-room scene, it reconstructs an incorrect global layout. Our method more faithfully recovers furniture arrangement, architectural structure, and surface appearance.

\begin{figure*}[t]
    \centering
    \includegraphics[width=\textwidth]{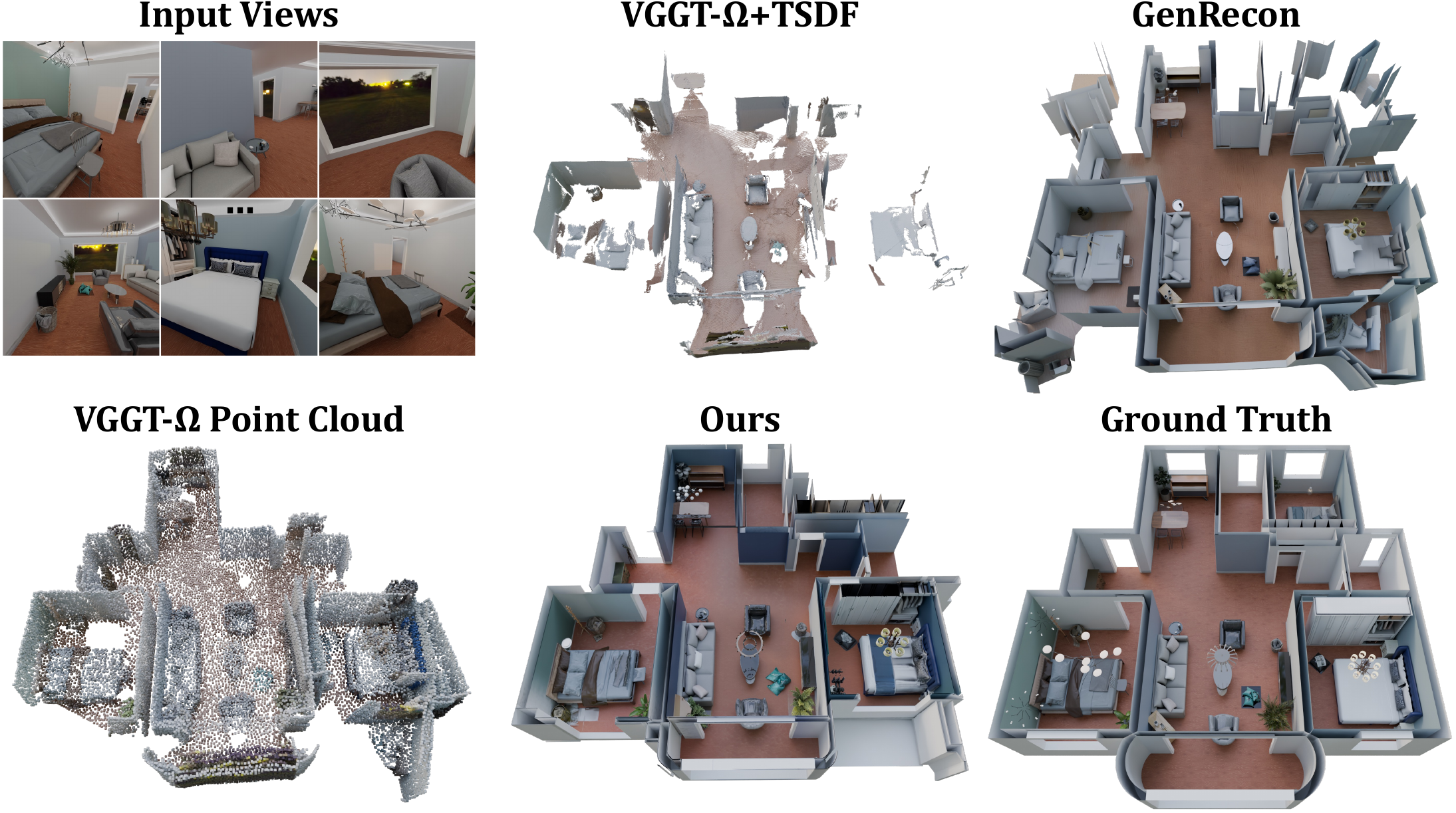}
    \vspace{-15pt}
    \caption{\textbf{Failure case.} Inaccurate reconstruction from VGGT-\(\Omega\) lead to missing geometry in unobserved regions. Nevertheless, \MethodName{} remains robust to substantial pose errors and still faithfully reconstructs the areas that are covered by the input views.}
    \label{fig:failure_case}
\end{figure*}

\subsection{Failure Cases and Limitations}

\Cref{fig:failure_case} illustrates a representative failure case on a multi-room scene. The 32 sampled input views (top-left) are visually rich, but parts of the scene are occluded or observed from only a handful of viewpoints. As a result, the VGGT-\(\Omega\) point-cloud scaffold (bottom-left) incorrectly estimates the poses of several views: some regions are either left unreconstructed or filtered out due to low confidence, while other views are projected to entirely wrong locations. 

Under this shared scaffold, the TSDF baseline can only reconstruct a small fraction of the
scene, while the generative baseline GenRecon directly reprojects image features using the erroneous poses, scattering misaligned projections across the scene and producing large fragmented artifacts as well as distorted objects.
\MethodName{} likewise suffers from missing geometry in unobserved regions
due to the incomplete input coverage. Nevertheless, it remains
robust to substantial pose errors and still faithfully reconstructs the areas
that are well covered by the input views, and is able to hallucinate plausible completions for the missing point-cloud regions based on sparse image cues.

\end{document}